\documentclass[letterpaper, 10 pt, conference]{ieeeconf}
\IEEEoverridecommandlockouts                              
\usepackage{verbatim}
\usepackage[table,xcdraw]{xcolor}

\usepackage{float}
\usepackage{cite}

\usepackage{hyperref}
\hypersetup{
   citecolor=black,
   colorlinks=true, 
   linktoc=all,     
   urlcolor=blue,
   linkcolor=black,  
}

\usepackage[pdftex]{graphicx}
\usepackage{multirow}

\usepackage{amsfonts}

\usepackage{amsmath}
\usepackage{algorithmic}
\usepackage[ruled, lined, linesnumbered, commentsnumbered, longend]{algorithm2e}

\usepackage{array}

 \title{\LARGE \bf An Efficient Approach to Generate Safe Drivable Space by LiDAR-Camera-HDmap Fusion}
\author{Minghao Ning$^{1}$, Ahmad Reza Alghooneh$^{1}$, Chen Sun$^{1}$, Ruihe Zhang $^{1}$ \\
Pouya Panahandeh$^{1}$, Steven Tuer$^{1}$, Ehsan Hashemi$^{2}$ and Amir Khajepour$^{1}$
\thanks{$^{1}$Minghao Ning, Ahmad Reza Alghooneh, Chen Sun, Ruihe Zhang, Pouya Panahandeh and Amir Khajepour are with the Mechanical and Mechatronics Eng. Department, University of Waterloo, 200 University Ave W, Waterloo, ON N2L3G1, Canada. e-mail:\{minghao.ning, aralghoo, chen.sun, r422zhang,  ppanahan, sktuer, a.khajepour\}@uwaterloo.ca).}
\thanks{$^{2}$Ehsan Hashemi is with the Mechanical Engineering Department, University of
Alberta, Alberta, T6G1H9, Canada (e-mail:ehashemi@ualberta.ca)}%
}

\begin{document}
\maketitle

\begin{abstract}
In this paper, we propose an accurate and robust perception module for Autonomous Vehicles (AVs) for drivable space extraction. Perception is crucial in autonomous driving, where many deep learning-based methods, while accurate on benchmark datasets, fail to generalize effectively, especially in diverse and unpredictable environments. Our work introduces a robust easy-to-generalize perception module that leverages LiDAR, camera, and HD map data fusion to deliver a safe and reliable drivable space in all weather conditions. We present an adaptive ground removal and curb detection method integrated with HD map data for enhanced obstacle detection reliability. Additionally, we propose an adaptive DBSCAN clustering algorithm optimized for precipitation noise, and a cost-effective LiDAR-camera frustum association that is resilient to calibration discrepancies. Our comprehensive drivable space representation incorporates all perception data, ensuring compatibility with vehicle dimensions and road regulations. This approach not only improves generalization and efficiency, but also significantly enhances safety in autonomous vehicle operations. Our approach is tested on a real dataset and its reliability is verified during the daily (including harsh snowy weather) operation of our autonomous shuttle, WATonoBus\cite{bhatt2023watonobus}.
\end{abstract}


\IEEEpeerreviewmaketitle

\section{Introduction}
\begin{figure*}[t]
    \centering
    \includegraphics[width=0.95\linewidth]{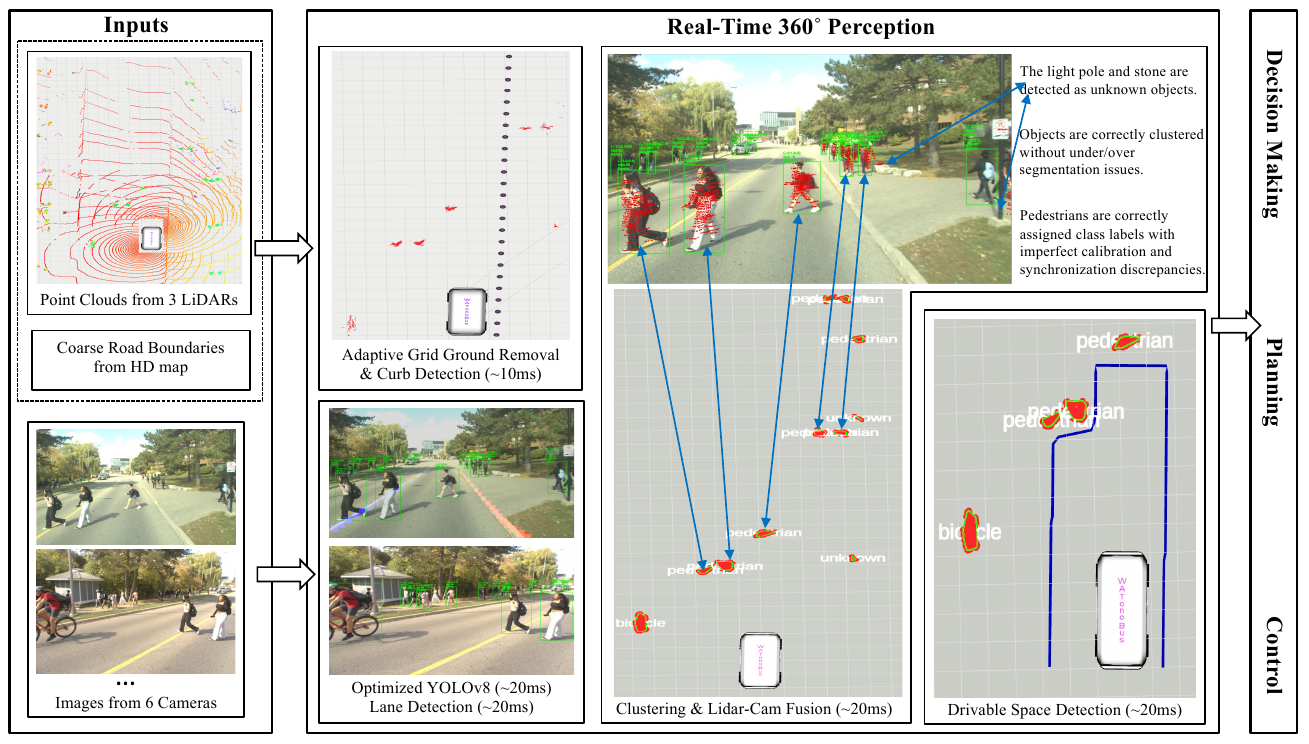}
    \vspace*{-0.2cm}
    \caption{Overview of the proposed method; multiple LiDAR point clouds, and in parallel, multiple camera images are received. Both undergo pre-processing, and then fused to come up with the reliable all-weather drivable space.   }
    \label{fig:framework}
\vspace*{-0.4cm}
\end{figure*}

Autonomous vehicles hold the promise of enhancing traffic safety and efficiency. Accurate and reliable perception is crucial for the safety of Autonomous Driving Systems (ADS), particularly in the context of Level 4 and Level 5 autonomous driving \cite{electronics11142162}. Current ADS employ a range of sensing modalities, including cameras, LiDAR, and radar. Despite this technological diversity, the limitations of individual sensors, deficiencies in perception algorithms, and shortcomings in the representation of perception results continue to impede the reliability of autonomous driving performance. Consequently, a 2018 survey by the U.S. Intelligent Transportation Systems Office reveals that 50$\%$ of consumers remain skeptical of self-driving technologies \cite{hassol2019understanding}. Therefore, the need to develop a reliable perception module remains essential.

One of the challenges facing the development of a reliable perception module is the performance during inclement weather. Degraded sensor resolution under severe environmental conditions can reduce sensors’ effectiveness in detecting objects accurately. In such conditions, the perception system may struggle to detect obstacles or other traffic participants, increasing the likelihood of accidents. 
Despite the cost-effectiveness of camera sensors, they are vulnerable to glare and reflections, which can obscure objects in the captured images \cite{sun2024real}. In addition, LiDARs also suffer from environmental impact, heavy data processing requirements, and short detection range drawback respectively \cite{9127855}. To overcome these individual sensor’s limitations, sensor fusion works are performed to further improve the reliability of perception performance. Specifically, point-cloud and image fusion techniques are heavily studied, these works employed different sensing modalities for detecting cars, pedestrians, and cyclists \cite{pang2020clocs, xie2020pi}. Nevertheless, these existing fusion-based studies have some shortcomings like dependence on accurate data calibration or association. Furthermore, raw data generally includes numerous redundant information, which can be challenging to implement for real-time processing.  

The dynamic traffic environment and diverse interacting objects often pose an Out of Distribution (OOD) challenge to the perception system. In real-world driving, there may be small animals, pedestrians in non-conventional attire, or specialized construction vehicles that are not included in the training set of the perception system \cite{sun2023toward}. To address the OOD challenge, robust learning techniques have been proposed to make neural networks more resilient to distribution shifts \cite{sun2023robust, sun2024extending}. Some research focuses on real-time detection of OOD cases to enable a safe transition to manual driving \cite{feng2021review}. However, validating robustness against real-world driving distributions is challenging. In recent years, researchers have explored approaches to bypass the OOD problem using more effective representations of perception results, such as occupancy grids or drivable space, to mitigate the impact of unknown classes. 


The contributions of this paper are highlighted as follows: 
\begin{itemize}

    \item An adaptive ground removal and curb detection method to accommodate OOD cases like snow piles, small animals and irregular shaped obstacles, to enhance safety. 

    \item An adaptive and efficient DBSCAN clustering robust to precipation noises e.g. snow and rain. The approach is able to maintain a consistently low false alarm rate in harsh weather through adjusting the DBSCAN parameters based on the scanning pattern of the LiDAR. 

    \item A cost based LiDAR-camera frustum association considering semantics and prior-based depth estimations from cameras, which is easily transferable and resilient to calibration errors and synchronization discrepancies. 

     \item A comprehensive drivable space representation, seamlessly integrating all perception data with ego vehicle states and HD map information, our solution ensures alignment with road laws and regulations, enhancing overall system integration and performance. 


    \item A scalable solution for AV perception. The method leverages high-definition (HD) map information, which is readily accessible for well-known and dynamic environments. Additionally, the approach is designed to be platform-agnostic, ensuring broad applicability across different AV systems.
\end{itemize}

The rest of the paper is organized as follows, in section \ref{sec:methods}, the overview of our method is presented, in section \ref{sec:experiment}, the details of our dataset, and experiments are presented, in section \ref{sec:results and discussion}, we present and discuss the results, and finally in section \ref{sec:conclusion} the impact of our work is concluded. 

\section{Methods}
\label{sec:methods}
Our method involves a series of precise steps to determine the drivable space for autonomous vehicles. Initially, the point clouds are cropped to the region of interest (ROI) defined by HD map data. Next, an adaptive ground removal method is proposed to segment points above the ground and curb detection are performed to non-ground points with the guide of HD map. Following this, an adaptive clustering method is used to group the non-ground points. Concurrently, two optimized neural networks are applied to the images to detect lanes and objects, which are then fused with the previously clustered point data. Finally, we utilize the data from obstacles, the right curb, and the centerline to precisely delineate the final drivable space. The framework of our approach is illustrated in Fig. \ref{fig:framework}.

\subsection{Inputs Pre-processing}


\subsubsection{Point Cloud Pre-processing, Consistent Concatenation}
To achieve 360\textdegree\ perception coverage, vehicles can are equipped with multiple LiDAR sensors, each producing a point cloud $\mathbb{P}_k = \{ P_k^1, P_k^2, \dots, P_k^{N_k} \}$ where $P_k^i = \{x, y, z, \text{intensity}, \text{ring}, \text{timestamp}\}$. These clouds are transformed into a unified coordinate system $G$ and concatenated into a comprehensive cloud $\mathbb{P}$. Due to motion during scans, timestamps of the points are utilized to align points accurately using odometry data.

\subsubsection{HD Map Utilization}
HD maps record the positions of the left and right road boundaries (curbs) and the localization information is used to transform the boundaries to vehicle's frame. While the maps provide road boundary data, the precision is not centimeter-level but sufficient for defining coarse road boundaries for the perception tasks. 


\subsection{Adaptive Grid Ground Removal and Curb Detection}
Initially, the entire point cloud is divided into grids based on the coarse road boundaries, where each grid undergoes a plane estimation to differentiate obstacle points from the ground. Non-ground points close to road boundaries are marked as candidate points for curb detection. For the region closest to the vehicle, it uses the height of the LiDAR sensor as the initial plane model, then iteratively improves the plane fitting. This estimation serves as the initial value for ground estimation in other regions. The output for the adaptive ground removal is a list of estimated ground plane models $\mathbb{F} = \{z=f_1(x,y), \cdots, z=f_M(x,y)\}$, where $M$ denotes the number of grids. This adaptive approach allows for dynamic adjustment of fitting thresholds, which are stricter closer to the vehicle to ensure the detection of small objects, thus enhancing safety. 

Using candidate curb points, the method then performs a robust polynomial fitting to model the curb's contour. This fitting is executed on a subset of the most relevant points, determined by their proximity to the road and their alignment with expected curb locations. The curb model is then used to refine the classification of points as either on-road or off-road, further enhancing the mapping of drivable and non-drivable spaces. The adaptation of this model allows for high accuracy in real-world scenarios, including those with varying curb heights and irregularities.

\subsection{Camera based Perception}
Cameras excel in capturing color and texture details but lack depth information. So, the images are used for 2D detection, i.e., detecting features on the image plane. 
Camera 2D detection, prominent in deep learning research, has more public datasets compared to camera 3D detection, as it's much easier to label 2D features on the image plane than 3D coordinates which often require complex setups for depth acquisition. Moreover, 2D detection methods are more consistent across different camera setups compared to 3D methods.
In our research, two state-of-the-art neural networks are optimized and then used to provide 2D object detection and 2D lane detection.
\subsubsection{YOLOv8}
YOLOv8 \cite{yolov8_ultralytics} is one of YOLO (You Only Look Once) series of real-time object detectors. It employs state-of-the-art backbone and neck architectures, making it achieving high accuracy with high inference speed. A customized dataset for outdoor autonomous driving usage has been created by selecting interesting objects from public datasets including COCO (Common Objects in Context) \cite{lin2015microsoft}, Argoverse \cite{Argoverse} and nuScenes \cite{nuscenes}. 
Then the YOLOv8 is trained on 90\% of the dataset, and evaluated on the rest 10\% of the dataset, and mAP50 of 0.708 has been achieved. Finally, the trained model is converted to TensorRT model for best inference speed. 
\subsubsection{UFLDv2}
UFLDv2 \cite{qin2022ultrav2} treats the lane detection task as an anchor-driven ordinal classification problem using global features, it can achieve both remarkable speed and accuracy. The output of the UFLDv2 is a set of lane marker points in the image plane, to obtain the lane marker position in the real world, the ground plane models $\mathbb{F}$ from the adaptive ground removal method are used to provide the road geometry information, then the 3D positions of the lane markers can be inferred using the inverse camera projection under the assumption that the markers are located on the estimated ground planes.

\subsection{Adaptive Clustering for All Weather Conditions}
Falling rain and snow can introduce amounts of noisy reflections to the LiDAR sensors.  Failing to de-noise point cloud may lead to false detection, and even trigger the emergency stop. Although clustering methods like DBSCAN \cite{DBSCAN} can somehow ease this by setting a smaller clustering distance threshold $\epsilon$ or a larger minimum number of points required to form a core region $\text{minPts}$. However, this will inevitably remove points from actual objects, and also introduce over-segmentation issue, where the points from the same object will be clustered into several groups. We improve this by proposing an adaptive DBSCAN clustering method that uses the LiDAR scanning pattern to adaptively adjust the $\epsilon$ and $\text{minPts}$.

Consider a plane object with height $h$ and width $\omega$ positioned at a distance of $s$ from the LiDAR, whose horizontal scanning resolution is $\Delta \varphi$ and vertical resolution is $\Delta \alpha$. For a more general case, the point cloud is usually first downsampled with a voxel size $\Delta d$ to reduce the processing time. Then the number of scan lines $N_{s}$ and the number of points in each line $N_{pl}$ falling to the plane object will be,
\begin{equation}
N_{s} = floor(\frac{h}{\max(s\Delta \alpha, \Delta d) }) 
\end{equation}
\begin{equation}
N_{pl} = floor(\frac{w}{\max(s\Delta \varphi, \Delta d) }) 
\end{equation}

Here, we propose an Adaptive-DBSCAN that dynamically adjusts the $\epsilon$ and $\text{minPts}$ based on the scanning pattern. In this method, the $\epsilon$ is adjusted to have a constant $N_{pl}$,
\begin{equation}
    N_{pl} = floor(\omega_{\min} / \Delta d) \ 
\end{equation}
where $\omega_{\min}$ is the minimum width of the objects in the environment.
Then the clustering distance threshold will be,
\begin{equation}
    \epsilon (s)= \max(\omega_{\min},\hspace{2.5pt}N_{pl} \Delta \varphi s)
\end{equation}

The number of scanning lines will be,
\begin{equation}
    N_{s} (s) = \max(1,\hspace{2.5pt}floor(\frac{h_{\min}}{\max(s\Delta \alpha,\hspace{2.5pt}\Delta d) }) )
\end{equation}
where $h_{\min}$ is the minimum height of the objects in the environment, and the $\text{minPts} = N_{s}N_{pl}$.

\subsection{Robust LiDAR Camera Fusion}
This part aims to fuse the object clustering results from the LiDAR and the 2D object detection results from the cameras to obtain clusters with semantic information. To achieve this, an association cost is calculated for each cluster and each 2D bounding box, and the Hungarian algorithm \cite{hungmatch} is used to find the best associations. The key to improving the LiDAR-camera fusion performance is the way to calculate the association cost. The 2D association cost based on the intersection over union (IOU) of the camera bounding box and the projected cluster points bounding box is a common choice due to its simplicity. However, it fails in handling distant objects where the bounding box is small and the intersection region could be zero due to imperfect calibration or object motion. We fix this issue by introducing the 3D information into the association cost.`1

\begin{figure}[tb]
    \centering
    \includegraphics[width=0.45\textwidth]{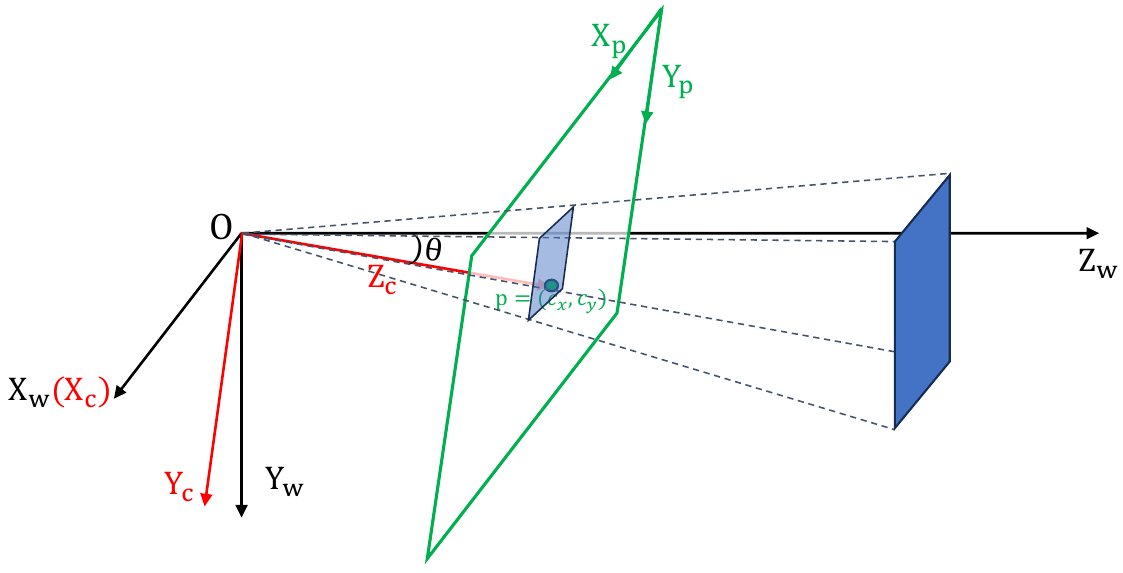}
    \vspace*{-0.2cm}
    \caption{Camera Projection}
    \label{fig:camera-projection}
    \vspace*{-0.4cm}
\end{figure}

A depth and its uncertainty estimation method using the semantic 2D bounding box is proposed. Considering a common camera placement scheme as shown in Fig. \ref{fig:camera-projection}, where there is a pitch angle $\theta$ between the camera coordinate plane $X_cOZ_c$ and the world coordinate plane $X_wOZ_w$ parallel to the ground surface. The projection of a point $P_w=(x_w,y_w,z_w)$ in the world coordinate to the pixel coordinate $p_p=(x_p, y_p)$ can be expressed as 
\begin{equation}
\begin{split}
    s \begin{bmatrix} x_p \\ y_p \\ 1 \end{bmatrix} &= K R_\theta \begin{bmatrix} x_w \\ y_w \\ z_w \end{bmatrix} \\
    &= \begin{bmatrix} f_x & 0 & c_x \\ 0 & f_y & c_y \\ 0 & 0 & 1 \end{bmatrix} \begin{bmatrix} 1 & 0 & 0 \\ 0 & \cos{\theta} & -\sin{\theta} \\ 0 & \sin{\theta} & \cos{\theta} \end{bmatrix} \begin{bmatrix} x_w \\ y_w \\ z_w \end{bmatrix}
\end{split}
\end{equation}

Based on this projection matrix, for an object having a fixed distance $z_w$, we have:
\begin{subequations}
\begin{equation}
  s = y_w \sin{\theta} + z_w \cos{\theta} 
\end{equation}    
\begin{equation}
  s \Delta x_p = f_x \Delta x_w
\end{equation}
\begin{equation}
  s \Delta y_p = ( (c_y-y_p) \sin{\theta} + f_y \cos{\theta} ) \Delta y_w
\end{equation}
\end{subequations}

Replace the $\Delta x_w$ and $\Delta y_w$ with $W_w$ and $H_w$, which denote the width and height of the object in the world coordinate. Similarly, replace the $\Delta x_p$ and $\Delta y_p$ with $W_p$ and $H_p$, representing the width and height of the 2D image bounding box of the object. The $y_w$ and $y_p$ will be the $y$-coordinate of the center of the object and bounding box, respectively. For an object standing on the ground, $y_w=H_w^{cam} - H_w/2$, where $H_w^{cam}$ denotes the height of the camera relative to the ground plane. So to calculate the final distance $z_w$ and its uncertainty $\sigma_z^2$, we need to calculate the depth estimation based on width and the height prior. Estimation based on width:
\begin{subequations}
    \begin{equation*}
        z_{w}^{\omega} = \frac{f_x W_w}{W_p \cos{\theta}} - y_w \tan{\theta}
    \end{equation*}
    \begin{equation*}
        \sigma_{z}^{\omega^2} = \frac{W_p^4 \sigma_{y_w}^2 \sin^2{\theta} + W_p^2 f_x^2 \sigma_{W_w}^2 + W_w^2 f_x^2 \sigma_{W_p}^2}{W_p^4 \cos^2{\theta}}
    \end{equation*}
    \begin{equation*}
        \sigma_{y}^{\omega^2} = \sigma_{H_w^{cam}}^2 + {\sigma_{H_w}^2}/{4}
    \end{equation*}
\end{subequations}

and the estimation based on height:
\begin{subequations}
    \begin{equation*}
        z_{w}^{h} = \frac{ ( (c_y-y_p) \sin{\theta} + f_y \cos{\theta} )H_w}{H_p \cos{\theta}} - y_w \tan{\theta}
    \end{equation*}
    \begin{align*}
        \sigma_{z}^{\omega^2} & =  \frac{1}{H_{p}^{4} \cos^{2}{\theta}}  \left[ H_{p}^{4} \sigma_{{H_{w}^{cam}}}^{2} \sin^{2}{\theta} \right. \\
    & + \frac{H_{p}^{2} \sigma_{H_w}^{2} \left(H_{p} \sin{\theta} + 2 f_{y} \cos{\theta} + 2 \left(c_{y} - y_{p}\right) \sin{\theta}\right)^{2}}{4} \\
    & + H_{w}^{2} \sigma_{H_p}^{2} \left(f_{y} \cos{\theta} + \left(c_{y} - y_{p}\right) \sin{\theta}^{2} \right]
    \end{align*}
\end{subequations}

Finally, the measurements are averaged based on the variances:
\begin{subequations}
\begin{equation}
  z_w = \frac{z_{w}^{h}\sigma_{z}^{\omega^2} + z_{w}^{\omega} \sigma_{z}^{h^2}}{\sigma_{z}^{\omega^2} + \sigma_{z}^{h^2}}
\end{equation}    
\begin{equation}
  \sigma_{z}^2 = \frac{\sigma_{z}^{h^2} \sigma_{z}^{\omega^2}}{\sigma_{z}^{\omega^2} + \sigma_{z}^{h^2}}
\end{equation}
\end{subequations}

This estimated depth information will be used as part of the association cost defined as,
\begin{equation}
    \pi_{i,j} = \delta(1-\text{IOU}_{i,j}) + (1-\delta)\Delta(\rho_{w}^{cam}, \rho_{w}^{Li})
\end{equation}
where, $\text{IOU}_{i,j}$ is calculated between the LiDAR projected bounding box, and the camera's, $\delta$ is the importance weight, $\Delta()$ takes the mahalanobis distance between the object position estimate from camera, $\rho_{w}^{cam}$, and from LiDAR, $\rho_{w}^{Li}$, and $\pi_{i,j}$ forms the Hungarian matrix $\Pi_{Li}^{cam}$, and can be solved with Hungarian matching algorithm in \cite{hungmatch}.

\subsection{Drivable Space Detection}
In constructing a comprehensive representation of drivable space, our method integrates the above lane and curb detection, objects with class information, and the HD map data. It accounts for ego vehicle dimensions and dynamics, object class information and road regulations, thereby facilitating a safe and lawful autonomous driving behavior.

Firstly, detected objects are projected onto the HD map to get the identification of objects positioned within the ego lane, the opposing lane, on sidewalks, or at crosswalks.
Subsequently, a binary occupancy map is constructed utilizing the positional data of objects, lanes, and curbs. This map shows occupied regions, which are further adjusted based on object class, as well as the dimensions and velocity of the ego vehicle to define a safety envelope for navigation.
For lateral expansion, the clearance required for each object is computed as $d_{class}+w_{ego}/2$, where $d_{class}$ is the predefined distance relative to the object’s class, and $w_{ego}$ is the width of the ego vehicle.
The longitudinal expansion incorporates an additional term representing the minimum braking distance, calculated as $v^2/(2a)$, where $v$ is the velocity of the ego vehicle and $a$ is the deceleration rate.
Vulnerable road users, such as pedestrians and cyclists, are allocated an increased clearance for safety reasons. A special case is applied to pedestrians on crosswalks, the lateral expansion is adjusted to span the entire road width so the ego vehicle will stop until the object walks off the crosswalk.

The refined binary occupancy grid is processed and simplified into a series of boundary points, which is the drivable space boundary. This is achieved by building a connection tree starting from the ego position that grows along the longitudinal direction and connects the safety region segments. This tree is searched and gives the final left-side and right-side drivable space boundary.

\section{Experiment}
\label{sec:experiment}
The performance of the proposed method is evaluated on the Waterloo all-weather autonomous shuttle (WATonoBus) \cite{bhatt2023watonobus}, which is equipped with 3 Robosense 32-line LiDARs, and 6 Basler cameras, a centimeter-level localization system Trimble APX-18 Land, and a computing unit NVIDIA Jetson AGX Orin. The bus travels along the 2.7 km Ringroad at the University of Waterloo, while high traffic volume and varying weather conditions make it a comprehensive test bed for autonomous driving.

The evaluation dataset contains 2 loops of sunny weather, 2 loops of heavy snow, 1 loop of light snow and a customized case where traffic cones are used to dynamically modify the drivable space region. The drivable space and the objects of interest are manually labeled.

\section{Results and Discussion}
\label{sec:results and discussion}

\begin{table}[htb]
\caption{Detection Evaluation}
\setlength{\tabcolsep}{2.5pt}
\begin{tabular}{ccc|cccc}
\hline
\multicolumn{3}{c|}{\textbf{Test Case}}                                                                                     & \textbf{HeavySnow}          & \textbf{LightSnow}          & \textbf{Sunny}              & \textbf{TrafficCone}        \\ \hline
\multicolumn{1}{c|}{}                              & \multicolumn{1}{c|}{}                                   & \textbf{CP}  & 28.85                       & 20.73                       & 8.63                        & 87.98                       \\
\multicolumn{1}{c|}{}                              & \multicolumn{1}{c|}{}                                   & \textbf{PP}  & 37.18                       & 13.55                       & 38.78                       & 88.54                       \\
\multicolumn{1}{c|}{}                              & \multicolumn{1}{c|}{\multirow{-3}{*}{\textbf{FAR5\%}}}  & \textbf{SSN} & 49.17                       & 38.37                       & 15.56                       & 95.36                       \\ \cline{2-7} 
\multicolumn{1}{c|}{}                              & \multicolumn{1}{c|}{}                                   & \textbf{CP}  & 18.89                       & 11.54                       & 7.18                        & 35.42                       \\
\multicolumn{1}{c|}{}                              & \multicolumn{1}{c|}{}                                   & \textbf{PP}  & 30.34                       & 11.03                       & 9.69                        & 70.78                       \\
\multicolumn{1}{c|}{\multirow{-6}{*}{\textbf{MR}}} & \multicolumn{1}{c|}{\multirow{-3}{*}{\textbf{FAR20\%}}} & \textbf{SSN} & 42.89                       & 21.02                       & 8.03                        & 82.85                       \\ \hline
\multicolumn{1}{c|}{\textbf{MR}}                   & \multicolumn{2}{c|}{\textbf{Our Method}}                               & {\color[HTML]{FE0000} 0.97} & {\color[HTML]{FE0000} 0.56} & {\color[HTML]{FE0000} 0.37} & {\color[HTML]{FE0000} 3.57} \\
\multicolumn{1}{c|}{\textbf{FAR}}                  & \multicolumn{2}{c|}{\textbf{Our Method}}                               & {\color[HTML]{FE0000} 1.24} & {\color[HTML]{FE0000} 0.76} & {\color[HTML]{FE0000} 0}    & {\color[HTML]{FE0000} 0}    \\ \hline
\end{tabular}
\label{table:detection-evaluation}
\end{table}
\begin{figure}[htb]
\vspace*{-0.5cm}
    \centering
    \includegraphics[width=0.37\textwidth]{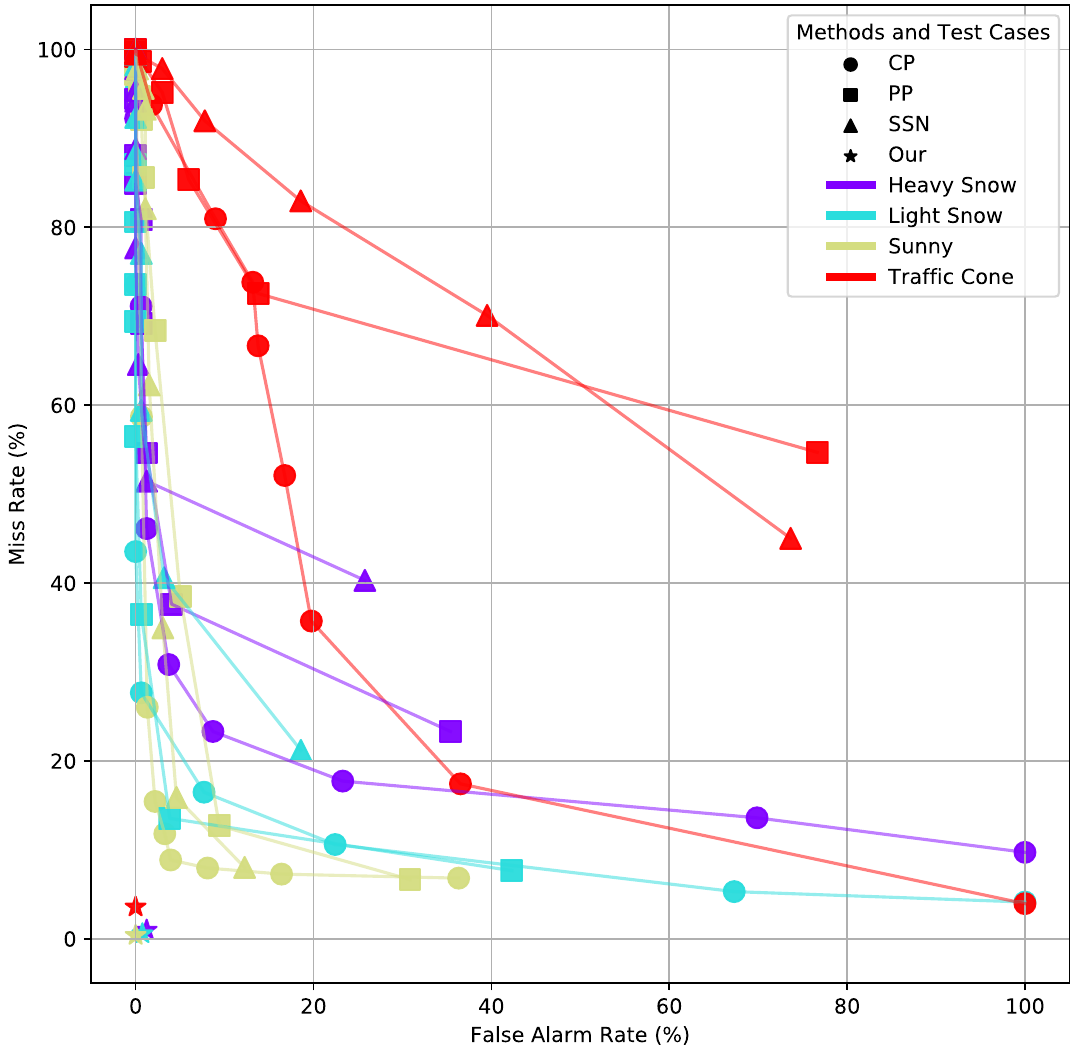}
    \vspace*{-0.3cm}
    \caption[Caption for LOF]{Detection performance comparison under various setups.}
    \label{fig:detection-evaluation}
    \vspace*{-0.2cm}
\end{figure}

The performance of the proposed adaptive ground removal and adaptive DBSCAN clustering are compared against methods such as CenterPoint\cite{yin2021center}, PointPillars\cite{lang2019pointpillars} and SSN\cite{zhu2020ssn}. Our results demonstrate a significant enhancement in both miss rate (MR) and false alarm rate (FAR), as shown in Fig. \ref{fig:detection-evaluation} and summarized in Table. \ref{table:detection-evaluation}.

\begin{figure}[htb]
    \centering
    \includegraphics[width=0.4\textwidth]{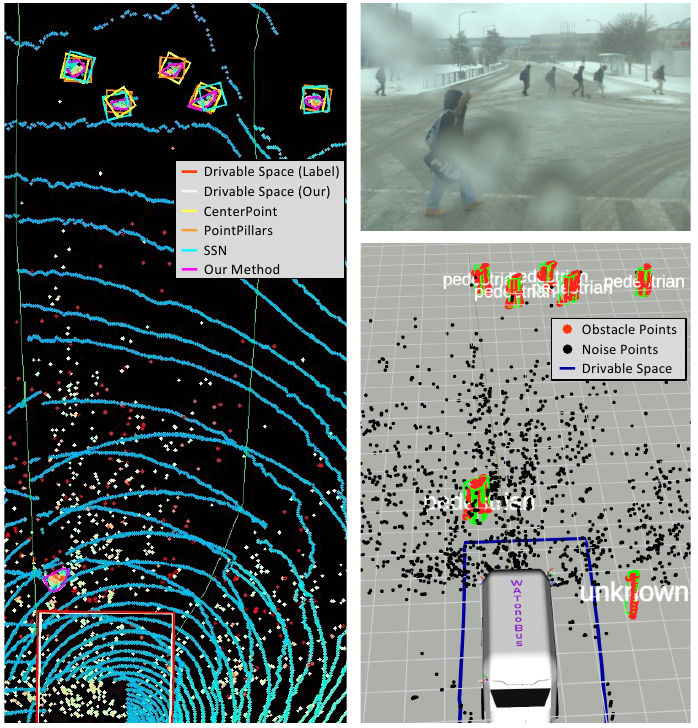}
    \vspace*{-0.2cm}
    \caption[Caption for LOF]{Detection performance in heavy snow conditions.\footnotemark}
    \label{fig:heavy-snow-case}
    \vspace*{-0.5cm}
\end{figure}

For the Heavy Snow and Light Snow cases, which present significant challenges to perception systems, our method achieved a remarkably low MR of 0.97\% and 0.56\% respectively, at a FAR of 1.24\% and 0.76\%. This is greatly better than the next best-performing method CP, which at a FAR of 5\%, exhibited MRs of 28.85\% and 20.73\% respectively. Such results indicate our method's superior capability under adverse weather conditions. 

Fig. \ref{fig:heavy-snow-case} presents a screenshot of the ego vehicle as it approaches an intersection during heavy weather conditions. The pre-trained neural networks failed to detect the pedestrian in front of the ego vehicle. This failure was due to the dense noise points surrounding the pedestrian, which disrupted the detection patterns learned from datasets collected under good weather conditions. In contrast, our method effectively distinguishes between noise and obstacle points. It successfully detects pedestrians and generates the drivable space, thereby enabling the bus to safely stop for pedestrians crossing in front.

In sunny conditions, our approach achieved an MR of 0.37\% with zero false alarms, thereby highlighting the remarkable accuracy of the proposed ground removal method in segmenting almost all obstacle points correctly. In contrast, deep-learning methods encountered false alarms, often caused by inaccurate orientation estimation for objects such as cars, and missed detections, primarily due to the sparsity of points for distant objects.

The Traffic Cone test case was particularly challenging due to the presence of dynamically placed small-sized traffic cones and the movement of people, where occlusions and sparse point data typically lead to elevated MRs. Our method exhibited an MR of 3.57\% without any false alarms, demonstrating adeptness in managing varying point density issues for the same objects at different distances. Deep-learning methods performed poorly in this scenario, attributed mainly to the point sparsity, especially concerning the traffic cones.

\footnotetext{A demonstration of real-time performance recording in heavy snow can be found \href{https://bit.ly/watonobus}{bit.ly/watonobus}}
\begin{table}[htb]
\caption{IOU Evaluation}
\begin{tabular}{ccccc}
\hline
\textbf{Test Case} & HeavySnow & LightSnow & Sunny & TrafficCone \\ \hline
\textbf{HDMap}     & 0.863     & 0.876     & 0.797 & 0.801       \\ \hline
\textbf{Our}       & 0.946     & 0.957     & 0.958 & 0.955       \\ \hline
\end{tabular}
\label{table:iou-evaluation}
\vspace*{-0.4cm}
\end{table}

The overall performance was further assessed based on the Intersection-Over-Union (IOU) between the computed drivable space and the human-labeled drivable space, as presented in Table \ref{table:iou-evaluation}. 
Utilizing only prior map information to construct the drivable space without considering the object detection, the HDMap served as a baseline. Our method demonstrated consistent superiority across various weather conditions.
In snowy conditions, with fewer objects present, the improvement is attributed to the accurate detection of the road boundaries when snow piles encroach upon the drivable space.
During sunny conditions, with more dynamic objects such as vehicles and pedestrians, our method achieved IOU of 0.958, significantly exceeding the HDMap baseline of 0.797.
In the challenging traffic cone scenario, where the drivable space is highly dynamically changing due to the varied placement of cones and pedestrians, our method still maintained a high IOU score of 0.955. 
Moreover, the correct detection of pedestrians on crosswalks resulted in a ``stop for the pedestrian" drivable space, which is more aligned with human-like behavior.

\section{Conclusion}
\label{sec:conclusion}
In conclusion, this paper presents several significant advancements in the perception systems of autonomous vehicles that collectively enhance driving safety and system reliability, particularly under challenging environmental conditions. Our innovative adaptive ground removal and curb detection method not only addresses small and irregular obstacles, but also integrates seamlessly with HD map data to improve detection reliability. The adaptive and efficient DBSCAN clustering technique we developed maintains a low false alarm rate even in adverse weather conditions by dynamically adjusting parameters according to LiDAR scanning patterns. Furthermore, our cost-based LiDAR-camera frustum association method successfully mitigates the challenges posed by imperfect calibration and synchronization discrepancies, proving robust across various scenarios. Lastly, the comprehensive drivable space representation our system employs not only leverages all available perception data but also respects vehicle dimensions and road laws, ensuring that our approach is both practical and compliant with existing regulations. These contributions demonstrate a significant leap forward in the practical application of autonomous driving technologies, paving the way for safer and more reliable autonomous vehicle operations in diverse and unpredictable environments.
\section*{Acknowledgment}
\label{sec: thanks}
The authors would like to acknowledge the financial support of the Natural Sciences and Engineering Research Council of Canada and the Canadian Foundation of Innovation in this work.


\bibliographystyle{IEEEtran}
\bibliography{./bibtex/bib/IEEEexample.bib}

\end{document}